# A REVIEW OF SCHEMES FOR FINGERPRINT IMAGE QUALITY COMPUTATION

*Fernando Alonso-Fernandez, Julian Fierrez-Aguilar, Javier Ortega-Garcia*

Biometrics Research Lab.- ATVS, Escuela Politecnica Superior - Universidad Autonoma de Madrid
Avda. Francisco Tomas y Valiente, 11 - Campus de Cantoblanco - 28049 Madrid, Spain
email: {fernando.alonso, julian.fierrez, javier.ortega}@uam.es

## ABSTRACT

Fingerprint image quality affects heavily the performance of fingerprint recognition systems. This paper reviews existing approaches for fingerprint image quality computation. We also implement, test and compare a selection of them using the MCYT database including 9000 fingerprint images. Experimental results show that most of the algorithms behave similarly.

## 1. INTRODUCTION

Due to its permanence and uniqueness, fingerprints are widely used in many personal identification systems. Fingerprints are being increasingly used not only in forensic environments, but also in a large number of civilian applications such as access control or on-line identification [1].

The performance of a fingerprint recognition system is affected heavily by fingerprint image quality. Several factors determine the quality of a fingerprint image: skin conditions (e.g. dryness, wetness, dirtiness, temporary or permanent cuts and bruises), sensor conditions (e.g. dirtiness, noise, size), user cooperation, etc. Some of these factors cannot be avoided and some of them vary along time. Poor quality images result in spurious and missed features, thus degrading the performance of the overall system. Therefore, it is very important for a fingerprint recognition system to estimate the quality and validity of the captured fingerprint images. We can either reject the degraded images or adjust some of the steps of the recognition system based on the estimated quality.

Fingerprint quality is usually defined as a measure of the clarity of ridges and valleys and the "extractability" of the features used for identification such as minutiae, core and delta points, etc [2]. In good quality images, ridges and valleys flow smoothly in a locally constant direction [3].

In this work, we review the algorithms proposed for computing fingerprint image quality. We also implement, test and compare a selection of them using the MCYT database [4, 5].

The rest of the paper is organized as follows. We review existing algorithms for fingerprint image quality computation in Sect. 2. An experimental comparison between selected techniques is reported in Sect. 3. Conclusions are finally drawn in Sect. 4.

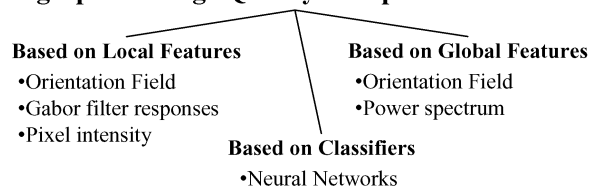

**Fig. 1**. A taxonomy of fingerprint image quality computation algorithms.

## 2. FINGERPRINT IMAGE QUALITY COMPUTATION

A taxonomy of existing approaches for fingerprint image quality computation is shown in Fig. 1. We can divide the existing approaches into $i$) those that use local features of the image; $ii$) those that use global features of the image; and $iii$) those that address the problem of quality assessment as a classification problem.

### 2.1. Based on local features

Methods that rely on local features [2, 3, 6-8] usually divide the image into non-overlapped square blocks and extract features from each block. Blocks are then classified into groups of different quality. A *local measure of quality* is finally generated. This local measure can be the percentage of blocks classified with "good" or "bad" quality, or an elaborated combination. Some methods assign a relative weight to each block based on its distance from the centroid of the fingerprint image, since blocks near the centroid are supposed to provide more reliable information [2, 8].

*2.1.1. Based on the orientation field*

This group of methods use the local angle information provided by the orientation field to compute several local features in each block. Hong et al. [3] modeled ridges and valleys as a sinusoidal-shaped wave along the direction normal to the local ridge orientation and extracted the amplitude, frequency and variance of the sinusoid. Based on these parameters, they classify the blocks as *recoverable* and *unrecoverable*. If the percentage of unrecoverable blocks exceeds a predefined threshold, the image is rejected. The method presented by Lim et al. [6] computes the following features in each block: orientation certainty level, ridge frequency, ridge thickness and ridge-to-valley thickness ratio. Blocks are then labeled as "good", "undetermined", "bad" or "blank" by thresholding the four local features. A local quality score $S_L$ is computed based on the total number of "good", "undetermined" and "bad" quality image blocks. Recently, Chen et al. [2] proposed a local quality index which measures the spatial coherence using the intensity gradient. The orientation coherence in each block is computed. A local quality score $Q_S$ is finally computed by averaging the coherence of each block, weighted by its distance to the centroid of the foreground.

*2.1.2. Based on Gabor filters*

Shen et al. [7] proposed a method based on Gabor features. Each block is filtered using a Gabor filter with $m$ different orientations. If a block has good quality (i.e. strong ridge orientation), one or several filter responses are larger than the others. In poor quality blocks or background blocks, the $m$ filter responses are similar. The standard deviation of the $m$ filter responses is then used to determine the quality of each block ("good" and "poor"). A quality index $QI$ of the whole image is finally computed as the percentage of foreground blocks marked as "good". If $QI$ is lower than a predefined threshold, the image is rejected. Poor quality images are additionally categorized as "smudged" or "dry".

*2.1.3. Based on pixel intensity*

The method described in [8] classifies blocks into "directional" and "non-directional" as follows. The sum of intensity differences $D_d(i,j)$ between a pixel $(i,j)$ and $l$ pixels selected along a line segment of orientation $d$ centered around $(i,j)$ is computed for $n$ different orientations. For each different orientation $d$, the histogram of $D_d(i,j)$ values is obtained for all pixels within a given foreground block. If only one of the $n$ histograms has a maximum value greater than a prominent threshold, the block is marked as "directional". Otherwise, the block is marked as "non-directional".

An overall quality score $Q$ is finally computed. A relative weight $w_i$ is assigned to each foreground block based on its distance to the centroid of the foreground. $Q$ is defined as $Q = \sum_D w_i / \sum_F w_i$ where $D$ is the set of directional blocks and $F$ is the set of foreground blocks. If $Q$ is lower than a threshold, the image is considered to be of poor quality. Measures of the smudginess and dryness of poor quality images are also defined.

**2.2. Based on global features**

Methods that rely on global features [2, 6] analyze the overall image and compute a *global measure of quality* based on the features extracted.

*2.2.1. Based on the orientation field*

Lim et al. [6] presented two features to analyze the global structure of a fingerprint image. Both of them use the local angle information provided by the orientation field, which is estimated in non-overlapping blocks. The first feature checks the continuity of the orientation field. Abrupt orientation changes between blocks are accumulated and mapped into a global orientation score $S_{GO}$. The second feature checks the uniformity of the frequency field [9]. This is done by computing the standard deviation of the ridge-to-valley thickness ratio and mapping it into a global score $S_{GR}$. Although ridge-to-valley thickness is not constant in fingerprint images in general, the separation of ridges and valleys in good quality images is more uniform than in low quality ones. Thus, large deviation indicates low image quality.

*2.2.2. Based on Power Spectrum*

Global structure is analyzed in [2] by computing the 2D Discrete Fourier Transform (DFT). For a fingerprint image, the ridge frequency value lies within a certain range. A region of interest (ROI) of the spectrum is defined as an annular region with radius ranging between the minimum and maximum typical ridge frequency values. As fingerprint image quality increases, the energy will be more concentrated in ring patterns within the ROI. The global quality index $Q_F$ defined in [2] is a measure of the energy concentration in ring-shaped regions of the ROI. For this purpose, a set of bandpass filters is constructed and the amount of energy in ring-shaped bands is computed. Good quality images will have the energy concentrated in few bands.

**2.3. Based on classifiers**

The method that uses classifiers [10] defines the quality measure as a degree of separation between the match and non-match distributions of a given fingerprint. This can be seen as a prediction of the matcher performance.

### 2.3.1. Based on neural networks

Tabassi et al. [10] presented a novel strategy for estimating fingerprint image quality. They first extract the fingerprint features used for identification and then compute the quality of each extracted feature to estimate the quality of the fingerprint image, which is defined as the degree of separation between the match and non-match distributions of a given fingerprint.

Let $s_m(x_i)$ be the similarity score of a genuine comparison (*match*) corresponding to the subject $i$, and $s_n(x_{ji})$, $i \neq j$ be the similarity score of an impostor comparison (*non-match*) between subject $i$ and impostor $j$. Quality $Q_N$ of a biometric sample $x_i$ is then defined as the prediction of

$$o(x_i) = \frac{s_m(x_i) - E[s_n(x_{ji})]}{\sigma(s_n(x_{ji}))} \quad (1)$$

where $E[.]$ is mathematical expectation and $\sigma(.)$ is standard deviation. Eq. 1 is a measure of separation between the *match* and the *non-match* distributions, which is supposed to be higher as image quality increases.

The prediction of $o(x_i)$ is done in two steps: $i)$ $v_i = L(x_i)$; and $ii)$ $Q_N = \tilde{o}(x_i) = I(v_i)$; where $L(.)$ computes a feature vector $v_i$ of $x_i$ and $I(.)$ maps the feature vector $v_i$ to a prediction $\tilde{o}(x_i)$ of $o(x_i)$ by using a neural network.

Feature vector $v_i$ contains the following parameters: $a$) number of foreground blocks; $b$) number of minutiae found in the fingerprint; $c$) number of minutiae that have quality value higher than 0.5, 0.6, 0.75, 0.8 and 0.9, respectively; and $d$) percentage of foreground blocks with quality equal to 1, 2, 3 and 4, respectively. All those values are provided by the MINDTCT package of NIST Fingerprint Image Software (NFIS) [11]. This method uses both local and global features to estimate the quality of a fingerprint.

## 3. EXPERIMENTS

In this work, we have implemented an tested some of the algorithms presented above using the existing fingerprint image database MCYT [4, 5]. In particular, 9000 fingerprint images from all the fingers of 75 subjects are considered (QMCYT subcorpus from now on). Fingerprints are acquired with an optical sensor, model UareU from Digital Persona, with a resolution of 500 dpi and a size of 400 pixels height and 256 pixels width. A subjective quality assessment $Q_M$ of this database was accomplished by a human expert. Each different fingerprint image has been assigned a subjective quality measure from 0 (lowest quality) to 9 (highest quality) based on factors like: captured area of the fingerprint, pressure, humidity, amount of dirt, and so on.

The algorithms tested in this work are as follows: $i)$ the combined quality measure $Q_C$ computed in [6] by linearly combining the scores $S_L$, $S_{GO}$ and $S_{GR}$ presented in

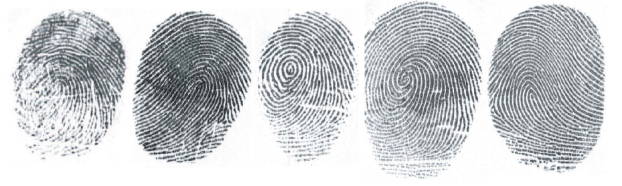

**Fig. 2**. Sample images extracted from the five quality subsets created using the manual quality measure $Q_M$. Images are arranged by increasing quality (on the left: lowest quality, subset 1; on the right: highest quality, subset 5).

Sects. 2.1.1 and 2.2.1; $ii)$ the algorithms presented in [2] based on local $Q_S$ (Sect. 2.1.1) and global features $Q_F$ (Sect. 2.2.2); and $iii)$ the method $Q_N$ based on neural networks proposed in [10] and described in Sect. 2.3.1. The quality measures $Q_C$, $Q_S$ and $Q_F$ lie in the range $[0, 1]$ whereas $Q_N \in \{1, 2, 3, 4, 5\}$. The selected methods are also compared with the subjective quality assessment $Q_M$ accomplished in QMCYT.

The above-mentioned quality measures have been computed for all QMCYT. In order to compare the selected methods, we have arranged the fingerprint images by increasing quality measure $Q_k$, $k \in \{M, N, C, S, F\}$. Then, 5 subsets $S_k^i$, $i \in \{1, 2, 3, 4, 5\}$, of equal size (1800 images per subset) are created. The first subset contains the 1800 images with the lowest quality measures, the second subset contains the next 1800 images with the lowest quality measures, and so on. Sample images extracted from the five quality subsets created using the manual quality measure $Q_M$ are shown in Fig. 2. The mean quality measure of each subset $S_k^i$ is then computed as $\tilde{Q}_k^i = \frac{1}{1800} \sum_{j \in S_k^i} Q_k^i(j) n(j)$ where $n(j)$ is the total number of images with quality measure $Q_k^i(j)$. Lastly, mean quality measures $\tilde{Q}_k^i$ are normalized to the $[0, 1]$ range as follows: $\hat{Q}_k^i = \left(\tilde{Q}_k^i - \tilde{Q}_k^1\right) / \left(\tilde{Q}_k^5 - \tilde{Q}_k^1\right)$ where $\hat{Q}_k^i$ is the normalized mean quality measure of $\tilde{Q}_k^i$.

In Fig. 3, we can see the normalized mean quality measures $\hat{Q}_k^i$ of each subset $S_k^i$, $i \in \{1, 2, 3, 4, 5\}$, for all the $k$ algorithms tested, $k \in \{M, N, C, S, F\}$. Maximum value, minimum value and standard deviation value of normalized individual quality measures of each subset are also depicted. It can be observed that that most of the algorithms result in similar behavior, assigning well-separated quality measures to different quality groups. Only the algorithm based on classifiers, $Q_N$, results in very different behavior, assigning the highest quality value to more than half of the database. It may be due to the low number of quality labels used by this algorithm [10].

Regarding to the algorithms that behave similarly, it can be observed that standard deviation is similar for quality groups 2 to 4. Only the method based on the subjective quality assessment $Q_M$ results in slightly higher deviation. This

5

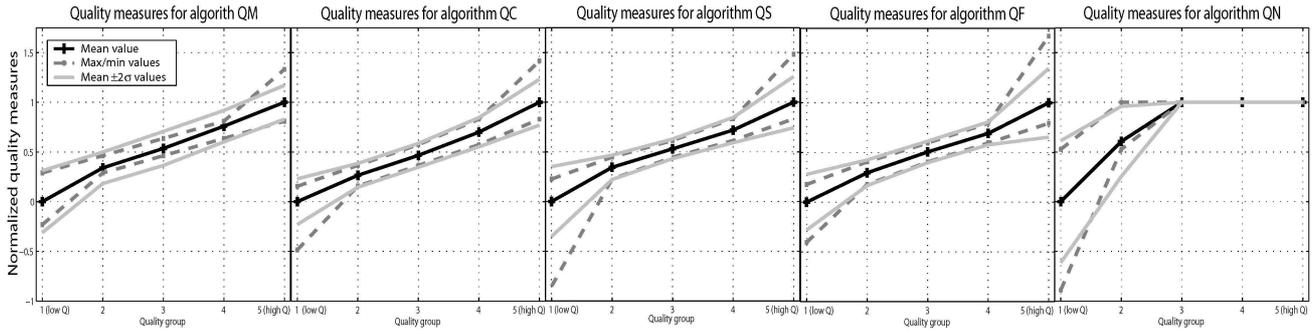

**Fig. 3**. Normalized mean quality measure $\hat{Q}_k^i$ of quality group $i \in \{1, 2, 3, 4, 5\}$, for all the $k$ algorithms tested (M=Manual, C=Combined local+global features [6], S=local spatial features [2], F=global frequency [2], N=classifier based on neural networks [10]). Maximum value, minimum value and standard deviation value of normalized quality measures of each quality group are also depicted.

is maybe due to the finite number of quality labels used. The other algorithms assign continuous quality measures within a certain range.

In addition, in most of the quality groups, normalized quality measures lie within a range of 2 times the standard deviation. Only quality groups 1 and 5 sometimes behave different, maybe to the presence of outliers (i.e., images with very low quality measure in group 1 and with very high quality measure in group 5, respectively).

## 4. CONCLUSIONS AND FUTURE RESEARCH

This paper reviews most of the existing algorithms proposed to compute the quality of a fingerprint image. They can be divided into $i$) those that use local features of the image; $ii$) those that use global features of the image; and $iii$) those that address the problem of quality assessment as a classification problem. We have implemented and tested a selection of them. They are compared with the subjective quality assessment accomplished in the existing QMCYT subcorpus. Experimental results show that most of the algorithms behave similarly, assigning well-separated quality measures to different quality groups. Only the algorithm based on classifiers [10] results in very different behavior. It may be due to the low number of quality labels used by this algorithm. Future work includes integrating the implemented quality estimation algorithms into a quality-based multimodal authentication system [12].

## Acknowledgments

This work has been supported by BioSecure European NoE and the TIC2003-08382-C05-01 project of the Spanish Ministry of Science and Technology. F. A.-F. and J. F.-A. thank Consejeria de Educacion de la Comunidad de Madrid and Fondo Social Europeo for supporting their PhD studies.